\documentclass[letterpaper]{article} 
\usepackage{aaai25}  
\usepackage{times}  
\usepackage{helvet}  
\usepackage{courier}  
\usepackage[hyphens]{url}  
\usepackage{graphicx} 
\usepackage{natbib}  
\usepackage{caption} 
\usepackage{subfiles}
\usepackage{algorithm}
\usepackage{algorithmic}

\usepackage{amsmath}
\usepackage{amssymb}
\usepackage{booktabs}
\usepackage{subcaption}
\usepackage{array}
\usepackage{textcomp}

\usepackage{verbatim}
\usepackage{cite}
\usepackage{multirow}
\usepackage{xcolor}

\usepackage{pdfpages}

\usepackage{newfloat}
\usepackage{listings}
\DeclareCaptionStyle{ruled}{labelfont=normalfont,labelsep=colon,strut=off} 
\floatstyle{ruled}
\newfloat{listing}{tb}{lst}{}
\floatname{listing}{Listing}
\title{Motif Guided Graph Transformer with Combinatorial Skeleton Prototype Learning for Skeleton-Based Person Re-Identification}
\author {
    Haocong Rao,
    Chunyan Miao\thanks{Corresponding author} 
}
\affiliations {
    College of Computing and Data Science, Nanyang Technological University (NTU), Singapore\\
    Joint NTU-UBC Research Centre of Excellence in Active Living for the Elderly (LILY), NTU, Singapore\\
     \{haocong001,ascymiao\}@ntu.edu.sg} 

\usepackage{bibentry}

\begin{document}

\maketitle

\begin{abstract}
Person re-identification (re-ID) via 3D skeleton data is a challenging task with significant value in many scenarios. Existing skeleton-based methods typically assume virtual motion relations between all joints, and adopt average joint or sequence representations for learning. However, they rarely explore key body structure and motion such as gait to focus on more important body joints or limbs, while lacking the ability to fully mine valuable spatial-temporal sub-patterns of skeletons to enhance model learning. This paper presents a generic Motif guided graph transformer with Combinatorial skeleton prototype learning (MoCos) that exploits \textit{structure-specific} and \textit{gait-related} body relations as well as combinatorial features of skeleton graphs to learn effective skeleton representations for person re-ID. In particular, motivated by the locality within joints' structure and the body-component collaboration in gait, we first propose the \textit{motif guided graph transformer (MGT)} that incorporates hierarchical structural motifs and gait collaborative motifs, which simultaneously focuses on multi-order local joint correlations and key cooperative body parts to enhance skeleton relation learning. Then, we devise the \textit{combinatorial skeleton prototype learning (CSP)} that leverages random spatial-temporal combinations of joint nodes and skeleton graphs to generate diverse \textit{sub-skeleton} and \textit{sub-tracklet} representations, which are contrasted with the most representative features (\textit{prototypes}) of each identity to learn class-related semantics and discriminative skeleton representations. Extensive experiments validate the superior performance of MoCos over existing state-of-the-art models. We further show its generality under RGB-estimated skeletons, different graph modeling, and unsupervised scenarios.
\end{abstract}

%
\begin{links}
    \link{Code}{https://github.com/Kali-Hac/MoCos}
\end{links}

\section{Introduction}
\label{sec:intro}

Person re-identification (re-ID) aims at matching and retrieving a person-of-interest from different views or scenes, which assumes an essential role in security authentication, smart surveillance, human tracking, and robotics 
\cite{vezzani2013people,ye2021deep}. Recent advancements in low-cost and accurate skeleton-tracking devices ($e.g.,$ Kinect \cite{shotton2011real-time}) have streamlined the collecting of 3D skeletons to make them as a popular and generic data modality for gait analysis and person re-ID 
\cite{liao2020model,rao2021self,rao2024hierarchical}. Compared with traditional person re-ID methods that rely on appearance features \cite{wang2016person}, 
skeleton-based models typically utilize body structural features and motion patterns of key body joints to identify different persons, which could possess many merits such as lighter inputs and models, better privacy protection ($e.g.,$ without using appearances or faces), and more robust performance under view, scale, and background variations \cite{han2017space}.
\begin{figure}
    \centering
    \scalebox{0.62}{
    \includegraphics{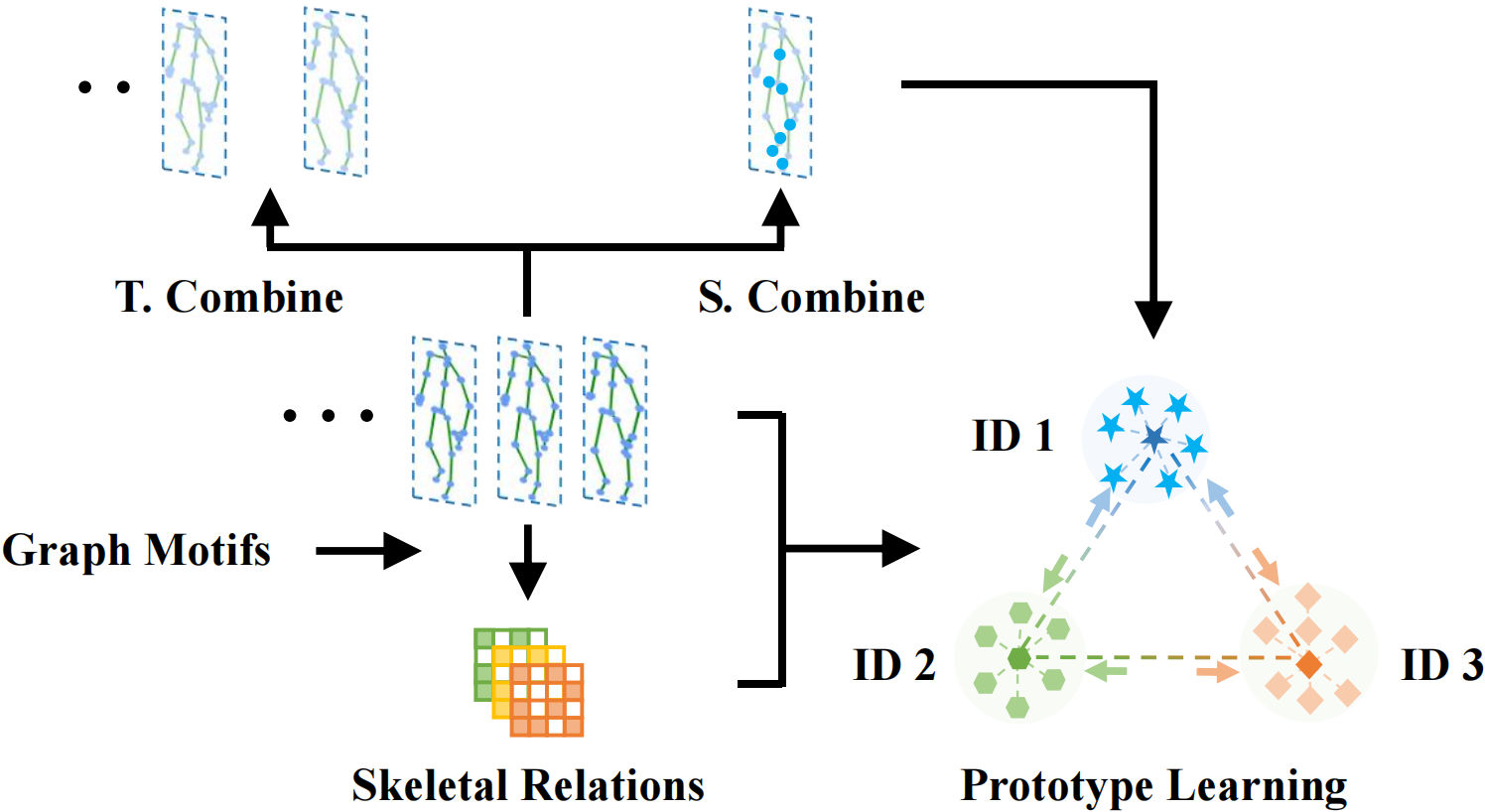}
    }
    \caption{Our approach exploits various graph motifs to enhance skeletal relation learning, and utilizes diverse spatial (S.) and temporal (T.) combinatorial skeleton features to perform skeleton prototype learning for person re-ID.}
    \label{first}
\end{figure}

Early skeleton-based methods \cite{andersson2015person} extract hand-crafted anthropometric and gait descriptors ($e.g.$, kinematic parameters) based on domain expertise, while they are often incapable of exploiting latent skeleton features beyond human cognition. Recent years have witnessed the great success of deep neural networks such as graph transformers for skeleton-based person re-ID \cite{liao2020model,rao2023transg}.
A common practice in these studies is to combine body-joint relation modeling and skeleton prototype learning ($e.g.,$ class feature clustering and contrasting) \cite{rao2022skeleton,rao2023transg}. However, most methods learn body joint or component relations with the assumption of virtual motion connections among \textit{all} joints \cite{rao2021multi,rao2023transg}, 
while they typically lack a specific \textit{focus} on key body joints or local body parts that are highly related to walking patterns ($e.g.,$ gait) to capture more discriminative features. 
On the other hand, existing works usually leverage \textit{average} skeleton or sequential features \cite{rao2022simmc} to perform representation learning, while they rarely harness different \textit{combinatorial} spatial or temporal patterns ($e.g.,$ sub-patterns) of key body joints, parts or skeletons to enhance the skeletal structure and motion learning. For example, different combinations of key joints such as hip and knee joints may characterize different structural features in walking patterns, while a combination of partial consecutive skeletons could contain key sub-patterns of a sequence, both of which can be utilized to mine more valuable skeleton features.

To address the aforementioned challenges, we propose a generic \textbf{Mo}tif guided graph transformer with \textbf{Co}mbinatorial \textbf{s}keleton prototype learning (MoCos) (illustrated in Fig. \ref{first}), which exploits different \textit{graph motifs} to guide body-joint relation learning in terms of key body structure and motion, and leverages different spatial-temporal feature combinations of both joints and skeletons to enhance skeleton graph representation learning for person re-ID.
In particular, motivated by the local correlations (referred to as \textit{locality}) within hierarchical body joints' structure, we first devise \textbf{\textit{hierarchical structural motifs (HSM)}}, which endow body joints with different semantic roles of connections, to specially focus on multi-order dependencies of body joints and their structural correlations to capture richer skeleton patterns. 
Then, considering that collaborative movements of upper and lower limbs usually contain unique ($e.g.,$ identity-specific) gait patterns \cite{murray1964walking}, we propose \textbf{\textit{gait collaborative motifs (GCM)}} that focus on both \textit{local} and \textit{global} motion relations of key limbs' joints to encourage the model to capture more salient gait features. By incorporating HSM and GCM into the joint relation learning, we devise the \textbf{\textit{motif guided graph transformer (MGT)}} to simultaneously capture key body relations from hierarchical local structure of joints and gait-related collaborative components for person re-ID.
Last, to exploit more valuable combinatorial patterns from skeletons and their sequences,
a \textbf{\textit{combinatorial skeleton prototype learning (CSP)}} approach is proposed to randomly mask body-joint nodes and skeleton graphs to generate spatial-temporal combinatorial graph features at both levels of \textit{sub-skeletons} and \textit{sub-tracklets}, which are utilized to contrast and learn the most representative skeleton graph features (referred to as \textit{prototypes}) of each identity.
CSP pulls different combinatorial skeleton graph representations closer to corresponding prototypes, and pushes them apart from other prototypes, so as to facilitate the model to learn distinguishing skeleton features and high-level class-related semantics for person re-ID.

Our main contributions can be summarized as follows:
\begin{itemize}
    \item We propose a generic MoCos paradigm that exploits diverse graph motifs and combinatorial skeleton features to learn effective representations from skeleton graphs for person re-ID. To the best of our knowledge, MoCos is the first exploration of structure-specific and gait-based graph motifs to enhance skeleton relation and prototype learning specifically for skeleton-based person re-ID. 
    \item We devise the motif guided graph transformer (MGT) by synergizing hierarchical structural motifs (HSM) and gait collaborative motifs (GCM) to guide body-joint relation learning, so as to capture more discriminative body structural and gait features within skeletons for person re-ID. 
    \item We propose the combinatorial skeleton prototype learning (CSP) that leverages combinatorial spatial-temporal graph features of joints (sub-skeletons) and skeletons (sub-tracklets) to learn more key skeleton patterns.
    \item Empirical evaluations on five public datasets validate that MoCos significantly outperforms existing state-of-the-art methods and can be effectively applied to different graph modeling, RGB-estimated or unsupervised scenarios.
\end{itemize}

\section{Related Works}
\label{sec:related}

\noindent\textbf{Skeleton-Based Person Re-Identification.}
Skeleton-based person re-ID focuses on the problem of matching and retrieving a certain person based on spatial and temporal representations of skeletal human body and gait \shortcite{rao20243Dsurvey,rao2024survey,fang2023you,fang2024not}.
Early-stage studies manually extract skeleton or body-joint descriptors in terms of anthropometric and gait attributes for person re-ID. \citeauthor{barbosa2012re} compute Euclidean distances between different joint pairs as descriptors, while they are further extended to 13 ($D_{13}$ \cite{munaro2014one}) and 16 skeleton descriptors ($D_{16}$ \cite{pala2019enhanced}) to perform person re-ID. 
Most recent methods \shortcite{liao2020model,rao2021self,rao2023transg,rao2024hierarchical} leverage deep learning models for skeleton sequence or skeleton graph representation learning.  
PoseGait \cite{liao2020model} is proposed to encode 3D pose features and joint-based motion descriptors (denoted as $D_{\text{PG}}$) for human recognition. \citeauthor{rao2020self} utilize an encoder-decoder model with attention mechanisms (AGE) to encode skeleton-based gait patterns, while its extension SGELA \cite{rao2021self} further enhances self-supervised skeleton semantic learning with diverse skeletal pretext tasks (e.g., time
series forecasting \cite{feng2024latent,zhichengsdformer}) and inter-sequence contrastive mechanisms for the person re-ID task.
\citeauthor{rao2022simmc} propose a masked contrastive learning framework (SimMC) to perform skeleton prototype learning with intra-sequence relation learning for person re-ID.
The multi-scale skeleton graphs are explored in \shortcite{rao2021multi,rao2021sm,rao2022skeleton} to learn body relations and patterns at various levels. In \cite{rao2023transg}, a skeleton graph transformer is devised to learn both skeleton and sequential graph features for person re-ID. A general skeleton feature re-ranking mechanism is proposed in \cite{rao2022revisiting} for skeleton-based person re-ID. Hi-MPC \cite{rao2024hierarchical} utilizes hierarchical prototype learning with a hard skeleton mining approach to learn discriminative skeleton features. Existing multi-modal person re-ID methods usually combine skeleton-based features with extra RGB or depth information ($e.g.,$ depth shape features based on point clouds \cite{munaro20143d,hasan2016long,wu2017robust}) to boost re-ID accuracy. 
For example, some works combine RGB images and skeleton data to learn auxiliary anthropometric attributes \citep{wang2020human}, body parts correlations \citep{lu2023exploring}, and clothing-invariant features \citep{nguyen2024attention} to enhance their performance.

\textbf{Graph Motifs.}
Motifs define different patterns of connections in graphs or networks via specifying the pattern-context nodes relevant to a target node of interest \cite{sankar2017motif}, which have been widely applied to many areas such as neuroscience and computer vision. \shortcite{sporns2004motifs,prvzulj2007biological,wen2022motif,gao2023hierarchical} 
A few recent works \cite{wen2022motif,wen2019graph} integrate motifs into graph convolutional networks (GCNs) to learn skeleton features from joints of interest and their context for action recognition.
As far as we know, this work is \textit{the first exploration} of structural and gait-based graph motifs with high-order semantic roles of joints \textit{specifically} for skeletal relation learning and person re-ID.

\begin{figure*}
    \centering
    \scalebox{0.58}{
    \includegraphics{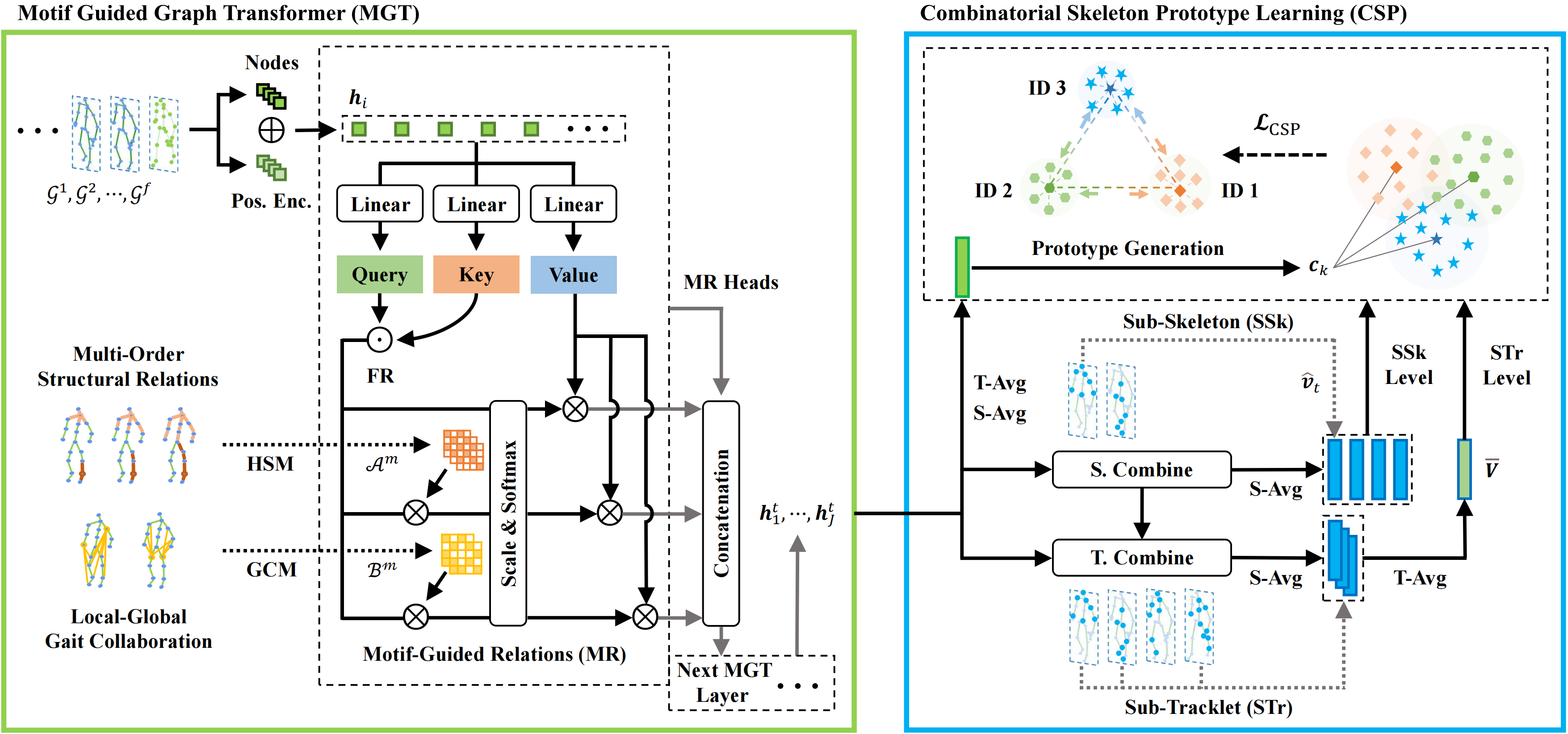}
    }
    \caption{Schematics of our approach: First, with position-encoded node representations for each skeleton graph $\mathcal{G}^{t}$, MGT 
    incorporates hierarchical structural motifs (HSM) and gait collaborative motifs (GCM) to perform body relation learning, which concurrently focuses on multi-order structural correlations and gait-related collaborative body parts to enhance skeleton pattern learning.
    Then, CSP temporally and spatially masks joints and graphs to generate combinatorial sub-skeleton (SSk) and sub-tracklet (STr) representations, which are contrasted with skeleton prototypes generated from same-identity spatially-temporally averaged (S-Avg and T-Avg) skeleton graph representations. We enhance the similarity of both SSk and STr level features to their corresponding prototypes, while maximizing their dissimilarity to other prototypes by optimizing $\mathcal{L}_{\mathrm{CSP}}$.}
    \label{model}
\end{figure*}

\section{Methodology}
\subsection{Preliminary}
\label{sec:preliminary}

\textbf{Problem Definition.} Suppose that a 3D skeleton sequence $\boldsymbol{X}\!=\!(\boldsymbol{x}_1,\cdots,\boldsymbol{x}_{f})\in \mathbb{R}^{f \times J \times 3}$, where $f$ denotes the number of skeletons in the sequence and $\boldsymbol{x}_{i}\in \mathbb{R}^{J \times 3}$ represents the $i^{th}$ skeleton with 3D coordinates of $J$ body joints.
 Each sequence $\boldsymbol{X}$ corresponds to an identity class $\text{y}\in \{1, \cdots, C\}$ and $C$ is the number of different identity classes. We denote the Training set, Probe set, and Gallery set as $\Phi_{\mathcal{T}}=\left\{\boldsymbol{X}^{\mathcal{T}}_{i}\right\}_{i=1}^{n_{1}}$, $\Phi_{\mathcal{P}}=\left\{\boldsymbol{X}^{\mathcal{P}}_{i}\right\}_{i=1}^{n_{2}}$, and $\Phi_{\mathcal{G}}=\left\{\boldsymbol{X}^{\mathcal{G}}_{i}\right\}_{i=1}^{n_{3}}$, which respectively contain $n_{1}$, $n_{2}$, and $n_{3}$ skeleton sequences of different persons collected from different scenes or views. 
 The model target is to encode skeleton sequences into effective representations, so that we can query the correct identity of each skeleton sequence representation (denoted as $\{\boldsymbol{V}^{\mathcal{P}}_i\}_{i=1}^{n_{2}}$) in the probe set via matching it with the sequence representations (denoted as $\{\boldsymbol{V}^{\mathcal{G}}_i\}_{i=1}^{n_{3}}$) in the gallery set.

\textbf{Skeleton Graph Construction.}
We construct skeleton graphs based on the physical connections of human body joints \cite{rao2023transg}: For the $t^{th}$ skeleton $\boldsymbol{x}_{t}$, we represent it as the graph $\mathcal{G}^t\left(\mathcal{V}^t, \mathcal{E}^t\right)$, which consists of $J$ nodes $\mathcal{V}^{t}=\{\boldsymbol{v}^{t}_{1}, \boldsymbol{v}^{t}_{2}, \cdots,\boldsymbol{v}^{t}_{J}\}$, $\boldsymbol{v}^{t}_{i}\in\mathbb{R}^{3}$, $i\in\{1,\cdots,J\}$ and edges $\mathcal{E}^{t}=\{e^{t}_{i,j}\ | \boldsymbol{v}^{t}_{i}, \boldsymbol{v}^{t}_{j}\!\in\!\mathcal{V}^{t}\}$, $e^{t}_{i,j}\in\mathbb{R}$. Here $\mathcal{E}^{t}$ denotes the set of connections and motion relations between different joints, and can be represented with an adjacent matrix $\mathbf{A}^{t} \in \mathbb{R}^{J \times J}$, initialized by the connections of adjacent body joints.

\subsection{Motif Guided Graph Transformer}
\label{MGT_sec}
Different body joints and parts of a pedestrian typically possess unique relations, such as structural relations between adjacent joints, and actional relations between non-adjacent parts, characterizing discriminative walking patterns \cite{murray1964walking,rao2021sm}. Existing methods typically perform global relation learning with the assumption of virtual motion relations among all joints \cite{rao2023transg}, while they rarely exploit local hierarchical structure of joints (defined as “\textit{locality}”) or key gait-related body components to capture richer valuable relations. To this end, we propose to endow body-joint nodes with different relational semantic roles (defined as “\textit{motifs}” for skeleton graphs),
and devise the \textbf{\textit{motif guided graph transformer (MGT)}} to simultaneously focus on their \textit{hierarchical structural relations} and \textit{gait collaborative relations} to learn effective skeleton graph representations for person re-ID.

\textbf{Graph Transformer (GT).} 
First, given a skeleton graph, its $J$ node representations are integrated with their positional encoding based on the graph adjacency matrix $\mathbf{A}^t$ \cite{rao2023transg}, which can be formulated as:
\begin{equation}
\boldsymbol{h}_i=\left(\mathbf{W}_1 \boldsymbol{v}_i+\boldsymbol{b}_1\right)+\left(\mathbf{W}_2 \boldsymbol{\lambda}_i+\boldsymbol{b}_2\right),
\end{equation}
where $\boldsymbol{h}_{i}\in\mathbb{R}^{D}$ represents the position-encoded representation of $i^{th}$ node, ${\lambda}_i\in\mathbb{R}^{K}$ denotes the $i$-node's positional encoding extracted from the $K$ smallest non-trivial eigenvectors of graph Laplacian
matrix following \cite{dwivedi2020generalization}, and
$\mathbf{W}_{1}\in \mathbb{R}^{D \times 3}, \mathbf{W}_{2}\in \mathbb{R}^{D \times K}, \boldsymbol{b}_{1}, \boldsymbol{b}_{2}\in \mathbb{R}^{D}$ are learnable parameters to map $i^{th}$ node $\boldsymbol{v}_{i}$ and corresponding positional encoding into feature spaces of the same dimension $D$.
Then, GT computes the \textit{preliminary} relation value of joints (referred to as “\textit{full relations (FR)}”) by 
\begin{equation}
\boldsymbol{R}_{i, j}^{k,l}=\operatorname{Softmax}_j\left( \frac{(\boldsymbol{Q}^{k, l} \boldsymbol{h}_{i}^{(l)}) \cdot (\boldsymbol{K}^{k, l} \boldsymbol{h}_{j}^{(l)})}{\sqrt{D_\text{k}}}\right).
\label{full_relation}
\end{equation}
In Eq. (\ref{full_relation}),  
$\boldsymbol{Q}^{k,l},\boldsymbol{K}^{k,l}\in\mathbb{R}^{D_\text{k}\times D}$ represent the learnable weight matrices for query and key transformations in the $k^{th}$ relation head of the $l^{th}$ GT layer, $\frac{1}{\sqrt{D_\text{k}}}$ is the scaling factor of dot-product similarity, and $\boldsymbol{R}_{i,j}^{k,l}$ denotes the $\operatorname{softmax}$-normalized relational value between the $i^{th}$ and $j^{th}$ joint captured by the $k^{th}$ relation head in the $l^{th}$ layer.

\subsubsection{Hierarchical Structural Motifs.}
\label{sec_HSM}
To guide the model to fully capture skeleton patterns from body joints' physical connections and the multi-level dependencies within their local hierarchical structure, we devise the \textit{hierarchical structural motifs (HSM)} to learn structural body relations from different-order neighbors of joint nodes. The focused body-joint relations of HSM can be represented as a matrix with
\begin{equation}
\mathcal{A}^{m}_{i, j}= \begin{cases}1 & \text { if } j \in  \bigcup\limits^{m}_{k=1} \mathcal{N}^{k}_{i}  \\ 0 & \text { otherwise } \end{cases},
\label{HSM_motifs}
\end{equation}
where $m\in\{1,2,3\}$, $\mathcal{A}^{m} \in \mathbb{R}^{J\times J}$ denote the $m$-order HSM matrix, $i\in \{1,2,\cdots,J\}$, and $\mathcal{N}^{k}_{i}$ represents the indices for $k$-order neighbors of the $i^{th}$ body-joint node ($i.e.,$  nodes with $k$-hop distance to the $i^{th}$ node). 
Intuitively, the $m$-order HSM $\mathcal{A}^{m}$ defines $R_{m}=2m+1$ semantic roles for all joint nodes: $\mathcal{A}^{1}$ contains $R_{1}=3$ roles, including a joint node itself, its parent node, and child node; 
$\mathcal{A}^{2}$ contains $R_{2}=5$ roles, including a joint node itself, its grandparent node, parent node, child node, and grandchild node, while $\mathcal{A}^{3}$ ($R_{3}=7$) further includes the roles of its great-grandparent node and great-grandchild node. 
Note that HSM does NOT require pre-defining the directions of node connections but views them as bi-directional to focus on the general hierarchical structure of joints. The maximum order is empirically set to 3 as the center joint of spine in most datasets has up to 3-hop neighbors \cite{li2021symbiotic}.
 By simultaneously focusing on relations of each body-joint node to its immediate and higher-level connected neighbors, 
HSM encourages the model to encode the inherent hierarchical structure ($e.g.,$ high-order locality) of joints' positions and motion to capture more valuable patterns of skeleton graphs.

\subsubsection{Gait Collaborative Motifs.} 
\label{sec_GCM}
Motivated by the gait property that different key body components ($e.g.,$ arms and legs) usually perform collaborative motion characterizing identity-specific patterns \cite{murray1964walking}, we propose the \textit{gait collaborative motifs (GCM)} to guide the model to learn more salient patterns from motion units of both upper and lower limbs (see Fig. \ref{model}). In particular, we regard each body joint in limbs as a basic motion unit, and focus on its local relations \textit{within the same limb} and global relations \textit{with other limbs} to facilitate the gait pattern learning. We define GCM with the focused body-joint relations as 
\begin{equation}
\mathcal{B}^{m}_{i, j}= \begin{cases}1 & \text { if } i \in  \mathcal{I}^{m}, \ j \in  \bigcup\limits^{2}_{k=1} \mathcal{I}^{k}, \ j\neq i \\ 0 & \text {otherwise}
\end{cases},
\label{limb_motifs}
\end{equation}
where $m\in\{1,2\}$, $\mathcal{B}^{1},\mathcal{B}^{2}\in \mathbb{R}^{J\times J}$ denote the GCM matrices for the upper and lower limbs, $\mathcal{I}^{1}$ and $\mathcal{I}^{2}$ represent the sets of indices for joint nodes in upper limbs ($e.g.,$ arms) and lower limbs ($e.g.,$ legs) respectively (visualized in Appendix I). Specifically, each GCM matrix defines $\hat{R}=3$ semantic roles for all joint nodes: $\mathcal{B}^{1}$ (or $\mathcal{B}^{2}$) contains the roles of a joint node in a upper (or lower) limb, its \textit{locally}-correlated sibling nodes ($i.e.$, nodes in the same limb), and its \textit{globally}-collaborative nodes in other limbs. In this way, GCM aims to focus on both local and global relation learning of limb motion units to encourage mining more unique cooperative skeleton patterns from their gait-related components.

By incorporating HSM and GCM into the relation learning process of GT, we devise the \textit{motif guided graph transformer (MGT)} to \textit{jointly} focus on hierarchical body-joint structure and gait-related components to capture more key skeleton patterns. 
In particular, MGT computes \textit{motif-guided relations (MR)} by updating Eq. (\ref{full_relation}) to (illustrated in Fig. \ref{model})
\begin{equation}
\boldsymbol{\hat{R}}_{i, j}^{k,l}=\operatorname{Softmax}_j\left( \frac{\mathcal{M}^{k}_{i, j}(\boldsymbol{Q}^{k, l} \boldsymbol{h}_{i}^{(l)}) \cdot (\boldsymbol{K}^{k, l} \boldsymbol{h}_{j}^{(l)})}{\sqrt{D_\text{k}}}\right),
\label{relation_motif}
\end{equation}
where 
\begin{equation}
\mathcal{M}^{k}_{i, j}= \begin{cases}\mathcal{A}^{k}_{i, j} & \text { if } k \in \{1, 2, 3\} \\ \mathcal{B}^{k-3}_{i, j} & \text { if }  k \in \{4, 5\} \\ 1 & \text { otherwise }  \end{cases}.
\label{motif_rule}
\end{equation}
In Eq. (\ref{relation_motif}) and (\ref{motif_rule}), 
$\boldsymbol{h}_{i}^{(l)} \in \mathbb{R}^{D}$ denotes the feature representation of the $i^{th}$ joint encoded by the $l^{th}$ MGT layer,
$\boldsymbol{\hat{R}}_{i,j}^{k,l}$ represents the relation value between the $i^{th}$ and $j^{th}$ joint computed by the $k^{th}$ MR head in the $l^{th}$ layer, $k\in\{1,2,\cdots,H\}$, and $H$ is the number of MR heads. MGT adopts multiple MR heads to jointly perform motif-guided and full relation learing, which are then aggregated into the each graph node representation with
\begin{equation}
	\boldsymbol{\hat{h}}_i^{(l)}=\boldsymbol{O}^{l} {\bigg\|}_{k=1}^{H}\left(\sum^{J}_{j=1} \boldsymbol{\hat{R}}_{i, j}^{k,l} \boldsymbol{V}^{k, l} \boldsymbol{h}_{j}^{(l)}\right),
\label{concat_head}
\end{equation}
where $\boldsymbol{V}^{k,l}\in\mathbb{R}^{D_\text{k}\times D}$ represent the learnable weight matrices for value transformation in the $k^{th}$ MR head of the $l^{th}$ MGT layer, $\boldsymbol{O}^{l} \in \mathbb{R}^{D \times D}$ is the parameter matrix for output transformation, ${\big\|}$ represents the concatenation operation, $\boldsymbol{\hat{h}}_i^{(l)}\in \mathbb{R}^{D}$ denotes the $i^{th}$ node representation that concatenates node features learned from different MR heads in the $l^{th}$ layer, 
For convenience, we use $\boldsymbol{h}^{t}_{i}$ to denote the final representation ($i.e.,$ concatenated node representation of the last MGT layer) of $i^{th}$ node in $t^{th}$ skeleton graph. 

By integrating different motifs (Eq. (\ref{HSM_motifs}), (\ref{limb_motifs})) into the relation computation (Eq. (\ref{relation_motif}), (\ref{motif_rule})), we encourage the model to focus on both multi-level structural relations and gait-related collaboration of key joints to capture richer effective patterns for person re-ID.  It is worth noting that the proposed multi-head MGT naturally generalizes the self-attention based GT \cite{rao2023transg} to local and global body relation learning using skeleton-specific motifs.
The proposed motifs can also be generally applied to non-graph models, unsupervised skeleton data, and different-scale skeleton representations.

\subsection{Combinatorial Skeleton Prototype Learning}
\label{CSP_sec}
To mine the most representative skeleton features of each identity for person re-ID, existing solutions \cite{rao2022skeleton,rao2023transg} typically \textit{average} spatial or temporal features of skeletons for prototype clustering and contrasting, while they rarely harness different \textit{combinatorial} spatial-temporal patterns of body joints, parts or skeletons to learn more effective representations.
 For example, a subset or dynamic combination of key body joints such as wrist, knee and foot joints (defined as “\textit{sub-skeleton representations}”) can depict different body structural features within gait, while different key segments of a skeletal walking tracklet (defined as “\textit{sub-tracklet representations}”) typically contain diverse sub-patterns \cite{zhang2020learning}, both of which can be exploited to learn more informative and unique features.
To this end, we propose the \textbf{\textit{combinatorial skeleton prototype learning (CSP)}} that leverages spatial-temporal combinatorial graph representations of sub-skeletons and sub-tracklets to jointly perform skeleton prototype learning.

Given the $t^{th}$ skeleton graph representation $(\boldsymbol{h}^{t}_{1},\cdots,\boldsymbol{h}^{t}_{J})$ containing $J$ spatial representations of body-joint nodes,
we utilize random masks to generate a subset of nodes to construct its spatial combinatorial representation ($i.e.,$ sub-skeleton representation) by  
\begin{equation}
\boldsymbol{\hat{v}}_{t}=\frac{1}{N_{S}}\sum^{J}_{j=1}{x}_{j}\boldsymbol{h}^{t}_{j},
\label{spatial_mask}
\end{equation}
where $\boldsymbol{\hat{v}}_{t}\in\mathbb{R}^{D}$ is the sub-skeleton representation of $t^{th}$ skeleton graph by randomly masking nodes, $x_{j}\in\{0, 1\}$ denotes the $j^{th}$ mask that is an independent and identically distributed Bernoulli random variable with the probability $p_{s}$ of being 0 ($i.e.$, ${x}_{j}\sim \operatorname{Bernoulli}(1-p_{s})$), and $N_{S}=\sum^{J}_{j=1}x_{j}$ represents the number ($i.e.,$ subset size) of unmasked node representations. Each unmasked node representation is assumed to be equally important and we average them to be the sub-skeleton graph representation.
In practice, the maximum number of masked node representations is $J-1$ ($i.e.,$ $N_{S}\geq1$) to avoid empty skeleton representation. Here we adopt Bernoulli distribution for combinatorial feature generation due to its simplicity and computational tractability \cite{boluki2020learnable}, while other probabilistic distributions can be also extended and applied to the proposed masking. 

Then, provided the spatial combinatorial representations $(\boldsymbol{\hat{v}}_{1}, \cdots, \boldsymbol{\hat{v}}_{f})$ of $f$ consecutive skeleton frames (defined as “\textit{a skeletal walking tracklet}”), we generate a random subset of the walking tracklet to yield the spatial-temporal combinatorial representation ($i.e.$, sub-tracklet representation) with
\begin{equation}
\boldsymbol{\overline{V}}=\frac{1}{N_{T}}\sum^{f}_{t=1}{m}_{t}\boldsymbol{\hat{v}}_{t},
\label{temporal_mask}
\end{equation}
where $\boldsymbol{\overline{V}}\in\mathbb{R}^{D}$ denotes the sub-tracklet representation that incorporates both spatial and temporal combinatorial features of a skeleton sequence, and
$m_{t}\in\{0, 1\}$ represents the $t^{th}$ random mask sampled from the Bernoulli distribution with the probability $p_{t}$ being 0. $N_{T}=\sum^{f}_{t=1}m_{t}$, $N_{T}\geq 1$ is the sequence length of sub-tracklet. Each sub-skeleton representation within the sub-tracklet is assigned with the same importance, and we average them as the final sub-tracklet representation. It is worth noting that a sub-tracklet contains sub-trajectory of partial body joints ($i.e.$, sub-skeletons), and can be regarded as a subset representation of sub-sequence trajectory.
In essence, the temporally-masked sequence in SimMC \cite{rao2022simmc} and average spatially-masked skeleton representations in TranSG \cite{rao2023transg} can be viewed as two special cases of proposed sub-tracklet representation by setting $p_{s}\!=\!0$ and $p_{t}\!=\!0$ respectively.

\begin{table*}[t]
\centering
\scalebox{0.7}{
\renewcommand\arraystretch{1.3}{
\setlength{\tabcolsep}{1.2mm}{
\begin{tabular}{ll|rrrr|rrrr|rrrr|rrrr|rrrr|rrrr}
\hline
\multicolumn{2}{l|}{\multirow{2}{*}{\textbf{Methods}}} & \multicolumn{4}{c|}{\textbf{BIWI-S}} & \multicolumn{4}{c|}{\textbf{BIWI-W}} & \multicolumn{4}{c|}{\textbf{KS20}} & \multicolumn{4}{c|}{\textbf{IAS-A}} & \multicolumn{4}{c|}{\textbf{IAS-B}} & \multicolumn{4}{c}{\textbf{KGBD}} \\ \cline{3-26} 
\multicolumn{2}{l|}{} & \textbf{mAP} & \textbf{R$_{1}$} & \textbf{R$_{5}$} & \textbf{R$_{10}$} & \textbf{mAP} & \textbf{R$_{1}$} & \textbf{R$_{5}$} & \textbf{R$_{10}$} & \textbf{mAP} & \textbf{R$_{1}$} & \textbf{R$_{5}$} & \textbf{R$_{10}$} & \textbf{mAP} & \textbf{R$_{1}$} & \textbf{R$_{5}$} & \textbf{R$_{10}$} & \textbf{mAP} & \textbf{R$_{1}$} & \textbf{R$_{5}$} & \textbf{R$_{10}$} & \textbf{mAP} & \textbf{R$_{1}$} & \textbf{R$_{5}$} & \textbf{R$_{10}$} \\ \hline
\multicolumn{1}{l|}{\multirow{3}{*}{\textbf{H.}}} & ${D_{\text{PG}}}$ \shortcite{liao2020model} & 6.7 & 18.5 & 45.4 & 63.8 & 8.7 & 6.5 & 15.5 & 20.3 & 11.3 & 35.2 & 61.5 & 70.5 & 11.0 & 16.4 & 39.5 & 53.4 & 10.6 & 16.0 & 41.2 & 57.3 & 2.1 & 30.0 & 49.1 & 58.1 \\
\multicolumn{1}{l|}{} & ${D_{13}}$ \shortcite{munaro2014one} & 13.1 & 28.3 & 53.1 & 65.9 & 17.2 & 14.2 & 20.6 & 23.7 & 18.9 & 39.4 & 71.7 & 81.7 & 24.5 & 40.0 & 58.7 & 67.6 & 23.7 & 43.7 & 68.6 & 76.7 & 1.9 & 17.0 & 34.4 & 44.2 \\
\multicolumn{1}{l|}{} & ${D_{16}}$ \shortcite{pala2019enhanced} & 16.7 & 32.6 & 55.7 & 68.3 & 18.8 & 17.0 & 25.3 & 29.6 & 24.0 & 51.7 & 77.1 & 86.9 & 25.2 & 42.7 & 62.9 & 70.7 & 24.5 & 44.5 & 69.1 & 80.2 & 4.0 & 31.2 & 50.9 & 59.8 \\ \hline
\multicolumn{1}{l|}{\multirow{5}{*}{\textbf{S.}}} & PoseGait \shortcite{liao2020model} & 9.9 & 14.0 & 40.7 & 56.7 & 11.1 & 8.8 & 23.0 & 31.2 & 23.5 & 49.4 & 80.9 & 90.2 & 17.5 & 28.4 & 55.7 & 69.2 & 20.8 & 28.9 & 51.6 & 62.9 & 13.9 & 50.6 & 67.0 & 72.6 \\
\multicolumn{1}{l|}{} & AGE \shortcite{rao2020self} & 8.9 & 25.1 & 43.1 & 61.6 & 12.6 & 11.7 & 21.4 & 27.3 & 8.9 & 43.2 & 70.1 & 80.0 & 13.4 & 31.1 & 54.8 & 67.4 & 12.8 & 31.1 & 52.3 & 64.2 & 0.9 & 2.9 & 5.6 & 7.5 \\
\multicolumn{1}{l|}{} & SGELA \shortcite{rao2021self} & 15.1 & 25.8 & 51.8 & 64.4 & 19.0 & 11.7 & 14.0 & 14.7 & 21.2 & 45.0 & 65.0 & 75.1 & 13.2 & 16.7 & 30.2 & 44.0 & 14.0 & 22.2 & 40.8 & 50.2 & 4.5 & 38.1 & 53.5 & 60.0 \\
\multicolumn{1}{l|}{} & SimMC \shortcite{rao2022simmc} & 12.3 & 41.7 & 66.6 & 76.8 & 19.9 & 24.5 & 36.7 & 44.5 & 22.3 & 66.4 & 80.7 & 87.0 & 18.7 & 44.8 & 65.3 & 72.9 & 22.9 & 46.3 & 68.1 & 77.0 & 11.7 & 54.9 & 66.2 & 70.6 \\
\multicolumn{1}{l|}{} & Hi-MPC \shortcite{rao2024hierarchical} & 17.4 & 47.5 & 70.3 & 78.6 & 22.6 & 27.3 & 40.3 & 48.8 & 22.0 & 69.6 & 83.5 & 87.1 & 23.2 & 45.6 & 67.3 & 75.4 & 25.3 & 48.2 & 70.2 & 77.8 & 10.2 & 56.9 & 70.2 & 75.1 \\ \hline
\multicolumn{1}{l|}{\multirow{5}{*}{\textbf{G.}}} & MG-SCR \shortcite{rao2021multi} & 7.6 & 20.1 & 46.9 & 64.1 & 11.9 & 10.8 & 20.3 & 29.4 & 10.4 & 46.3 & 75.4 & 84.0 & 14.1 & 36.4 & 59.6 & 69.5 & 12.9 & 32.4 & 56.5 & 69.4 & 6.9 & 44.0 & 58.7 & 64.6 \\
\multicolumn{1}{l|}{} & SM-SGE \shortcite{rao2021sm} & 10.1 & 31.3 & 56.3 & 69.1 & 15.2 & 13.2 & 25.8 & 33.5 & 9.5 & 45.9 & 71.9 & 81.2 & 13.6 & 34.0 & 60.5 & 71.6 & 13.3 & 38.9 & 64.1 & 75.8 & 4.4 & 38.2 & 54.2 & 60.7 \\
\multicolumn{1}{l|}{} & SPC-MGR \shortcite{rao2022skeleton} & 16.0 & 34.1 & 57.3 & 69.8 & 19.4 & 18.9 & 31.5 & 40.5 & 21.7 & 59.0 & 79.0 & 86.2 & 24.2 & 41.9 & 66.3 & 75.6 & 24.1 & 43.3 & 68.4 & 79.4 & 6.9 & 40.8 & 57.5 & 65.0 \\
\multicolumn{1}{l|}{} & ST-GCN \shortcite{yan2018spatial} & 28.5 & 61.6 & 78.2 & 89.5 & 28.2 & 32.9 & 47.6 & 54.8 & 40.1 & 60.4 & 79.9 & 84.6 & 34.0 & 41.6 & 60.6 & 68.2 & 28.1 & 49.1 & 68.1 & 76.3 & 21.1 & 57.7 & 71.6 & 77.2 \\
\multicolumn{1}{l|}{} & TranSG \shortcite{rao2023transg} & 30.1 & 68.7 & 86.5 & 91.8 & 26.9 & 32.7 & 44.9 & 52.2 & 46.2 & 73.6 & 86.3 & \textbf{90.2} & 32.8 & 49.2 & 68.5 & 76.2 & 39.4 & 59.1 & 77.0 & 87.0 & 20.2 & 59.0 & 73.1 & 78.2 \\
\multicolumn{1}{l|}{} & \textbf{MoCos (Ours)} & \textbf{32.1} & \textbf{72.0} & \textbf{89.5} & \textbf{93.0} & \textbf{30.5} & \textbf{36.0} & \textbf{49.2} & \textbf{57.0} & \textbf{50.8} & \textbf{76.0} & \textbf{87.3} & \textbf{90.2} & \textbf{35.8} & \textbf{51.9} & \textbf{69.4} & \textbf{77.5} & \textbf{45.5} & \textbf{61.5} & \textbf{79.1} & \textbf{87.8} & \textbf{26.1} & \textbf{62.0} & \textbf{75.2} & \textbf{79.6} \\ \hline
\end{tabular}
}
}
}
\caption{Person re-ID performance comparison with state-of-the-art \textbf{H}and-crafted methods (\textbf{H.}), \textbf{S}equence representation learning methods (\textbf{S.}), and \textbf{G}raph-based methods (\textbf{G.}).
\textbf{Bold numbers} denote the best performance results among all methods. }
\label{formal_results}
\end{table*}

To exploit graph representations of both sub-skeletons and sub-tracklets to learn the most discriminative skeleton graph features (defined as “\textit{prototypes}”) of each person and high-level semantics ($e.g.,$ identity-associated patterns), we propose the combinatorial skeleton prototype (CSP) loss as
\begin{equation}
\mathcal{L}_{\mathrm{CSP}}=\lambda \mathcal{L}^{str}_{\mathrm{CSP}} + (1-\lambda) \mathcal{L}^{ssk}_{\mathrm{CSP}},
\label{CSP_loss}
\end{equation}
where
\begin{equation}
\mathcal{L}^{str}_{\mathrm{CSP}}=\frac{1}{n_{1}} \sum_{i=1}^{n_{1}}-\log \frac{\exp \left(\boldsymbol{\overline{V}}_{i} \cdot \boldsymbol{\overline{c}} / \tau_{1}\right)}{\sum_{k=1}^{C} \exp \left(\boldsymbol{\overline{V}}_{i} \cdot \boldsymbol{c}_{k} / \tau_{1}\right)},
\label{CSP_loss_seq}
\end{equation}
\begin{equation}
\scalebox{0.91}{
$
\begin{aligned}
    \mathcal{L}_{\mathrm{CSP}}^{ssk}=\frac{1}{f n_1} \sum_{i=1}^{n_{1}}\sum_{t=1}^{f}-\log \frac{\exp \left(\mathcal{F}_1\left(\boldsymbol{\hat{v}}_{t}^{i}\right) \cdot \mathcal{F}_2\left(\boldsymbol{\hat{c}}\right) / \tau_2\right)}{\sum_{k=1}^C \exp \left(\mathcal{F}_1\left(\boldsymbol{\hat{v}}_{t}^{i}\right) \cdot \mathcal{F}_2\left(\boldsymbol{c}_{k}\right) / \tau_2\right)},
\label{CSP_loss_ske}
\end{aligned}
$
}
\end{equation}
\begin{equation}
\boldsymbol{c}_{k}=\frac{1}{u_k}\sum^{}_{\text{y}_{j}=k}\boldsymbol{V}_{j}.
\label{proto_compute}
\end{equation}
The proposed CSP loss in Eq. (\ref{CSP_loss}) combines both \textit{sub-tracklet-level} ($\mathcal{L}^{str}_{\mathrm{CSP}}$) and \textit{sub-skeleton-level} combinatorial prototype loss ($\mathcal{L}^{ssk}_{\mathrm{CSP}}$) with the fusion coefficient $\lambda$. 
In Eq. (\ref{CSP_loss_seq}), (\ref{CSP_loss_ske}) and (\ref{proto_compute}), $n_{1}$ is the number of training skeleton sequences, $\boldsymbol{\hat{v}}_{t}^{i}$ and $\boldsymbol{\overline{V}}_{i}$ denote the sub-skeleton representation of the $t^{th}$ skeleton (see Eq. (\ref{spatial_mask})) and the sub-tracklet representation of the $i^{th}$ skeleton sequence (see Eq. (\ref{temporal_mask})). $\boldsymbol{\overline{c}}$ and $\boldsymbol{\hat{c}}$ correspond to their prototypes ($i.e.,$ class feature centroids) generated by averaging all sequence representations of the same identity  (see Eq. (\ref{proto_compute})), $\boldsymbol{c}_{k}$ is the skeleton prototype of $k^{th}$ class, $u_{k}$ denotes the number of skeleton sequence representations $\boldsymbol{V}_{j}$ with the class label $\text{y}_{j}=k$, and $\tau_{1}$, $\tau_{2}$ represent temperatures for contrastive learning. $\mathcal{F}_{1}(\cdot)$ and $\mathcal{F}_{2}(\cdot)$ are learnable projections to transform sequence-level prototypes and sub-skeleton-level features into the same feature space and integrate related features for contrastive learning. 
$\mathcal{L}_{\mathrm{CSP}}$ can be viewed as a generalized skeleton prototype loss that incorporates \textit{joint-level} motif-guided relation learning, \textit{sub-skeleton-level} and \textit{sub-tracklet-level} prototype contrasting to enhance spatial-temporal skeleton pattern learning, which can be theoretically modeled as a generalized Expectation-Maximization (EM) solution (see Appendix II). 




\section{Experiments}
\label{experiments}
\subsection{Experimental Setups}
\textbf{Datasets.} Four skeleton-based person re-ID benchmark datasets are used to evaluate our approach, including \textit{IAS} \cite{munaro2014feature}, \textit{KS20} \cite{nambiar2017context}, \textit{BIWI} \cite{munaro2014one}, \textit{KGBD} \cite{andersson2015person}, which contain 11, 20, 50, and 164 different persons. 
The generality of MoCos is also validated on a large-scale multi-view gait dataset \textit{CASIA-B} \cite{yu2006framework} with RGB-estimated skeleton data of 124 individuals under three conditions (Normal (N), Bags (B), Clothes (C)). The commonly-used standard probe and gallery settings \cite{rao2023transg} are adopted for a fair comparison.


\textbf{Implementation Details.} The skeletons in KGBD, IAS, and BIWI contain $J=20$ body joints, while KS20 and CASIA-B (RGB-estimated skeletons) contain $J=25$ and $J=14$ joints, respectively. 
For a fair comparison, we follow existing methods \cite{rao2024hierarchical} to set the sequence length to $f=6$ for IAS, KS20, BIWI, KGBD and $f=40$ for the RGB-estimated skeleton data in CASIA-B. 
We set the embedding size to $D=128$ for each node representation, and empirically employ 2 MGT layers with $H=8$ relation heads and $D_{\text{k}}=16$ for each layer.
The probability for spatial or temporal masking  of CSP is empirically set for different datasets: $p_{s}=0.25$, $p_{t}=0.25$ for IAS, BIWI, KS20, and $p_{s}=0.5$, $p_{t}=0.25$ for KGBD.
We use fusion coefficient $\lambda=0.9$ for BIWI-W, KGBD, KS20, $\lambda=0.25$ for BIWI-S, $\lambda=0.75$ for IAS-A and IAS-B.
We set the learning rate to $3.5\times 10^{-4}$ and use an Adam optimizer with batch size 256 for model training on all datasets. More technical details are provided in the appendices.

\textbf{Evaluation Metrics.}
Cumulative matching characteristics curve is computed and we report Rank-1, Rank-5, Rank-10 accuracy (R$_{1}$, R$_{5}$, R$_{10}$), and Mean Average Precision (mAP) \cite{zheng2015scalable} to evaluate model performance.

\subsection{Comparison with State-of-the-Art Methods}

Our approach is compared with state-of-the-art hand-crafted methods, sequence learning methods, and graph-based methods on BIWI, KS20, IAS, KGBD in Table \ref{formal_results}.


\textbf{Comparison with Graph-based Methods:}
As shown in Table \ref{formal_results}, 
the proposed MoCos significantly outperforms existing state-of-the-art graph-based methods (SPC-MGR \cite{rao2022skeleton}, MG-SCR \cite{rao2021multi}, SM-SGE \cite{rao2021sm}) with an improvement of $11.1$-$41.3\%$ for mAP and $10.0$-$51.9\%$ for Rank-1 accuracy on different benchmark datasets. Unlike these methods that resort to multi-scale graph modeling and multi-stage relation learning, 
our approach can utilize simpler \textit{single-level} graph representations with motif guided \textit{concurrent} relation learning (MGT) to more effectively capture distinguishing skeleton features for person re-ID.
In contrast to the latest skeleton graph model TranSG \cite{rao2023transg} employing na\"ive GT, our model using MGT also consistently achieves better performance in terms of mAP ($2.0$-$6.1\%$), Rank-1 ($2.4$-$3.3\%$), Rank-5 accuracy ($0.9$-$4.3\%$), and Rank-10 accuracy ($0.0$-$4.8\%$) on all datasets. This demonstrates the higher efficacy of MoCos incorporating joint-level motif-guided relation learning and different-level prototypical contrast (CSP) to learn richer unique skeleton features for person re-ID.
We will also discuss its generality under diverse graph modeling and different unsupervised paradigms in the next section.

\textbf{Comparison with Hand-crafted and Sequence Learning Methods:}
Compared with methods that rely on hand-crafted pose features ($D_{\text{PG}}$ \cite{liao2020model}) or anthropometric attributes ($D_{13}$ \cite{munaro2014one}, $D_{16}$ \cite{pala2019enhanced}), our approach achieves superior performance by a marked margin of up to $53.5\%$ Rank-1 accuracy and  $39.5\%$ mAP on different benchmarks.
Moreover, MoCos also obtains significantly higher performance than latest skeleton sequence contrastive models SimMC \cite{rao2022simmc} and Hi-MPC \cite{rao2024hierarchical} that utilize temporally-masked or hierarchical skeleton representations. This demonstrates the stronger effectiveness of our skeleton graph contrastive model (CSP) that combines both sub-skeleton and sub-tracklet level spatial-temporal features to capture more recognizable patterns for person re-ID.



\begin{table}[t]
\scalebox{0.7}{
\renewcommand\arraystretch{1.3}{
\setlength{\tabcolsep}{1.15mm}{
\begin{tabular}{cccccrrrrrrrr}
\hline
\multirow{2}{*}{\textbf{ID}} & \multirow{2}{*}{\textbf{GT}} & \multirow{2}{*}{\textbf{HSM}} & \multirow{2}{*}{\textbf{GCM}} & \multirow{2}{*}{\textbf{CSP}} & \multicolumn{2}{c}{\textbf{BIWI-S}} & \multicolumn{2}{c}{\textbf{BIWI-W}} & \multicolumn{2}{c}{\textbf{KS20}} & \multicolumn{2}{c}{\textbf{KGBD}} \\
 &  &  &  &  & \textbf{R$_{1}$} & \textbf{mAP} & \textbf{R$_{1}$} & \textbf{mAP} & \textbf{R$_{1}$} & \textbf{mAP} & \textbf{R$_{1}$} & \textbf{mAP} \\ \hline
\textbf{1} &  &  &  &  & 38.1 & 11.3 & 21.2 & 18.3 & 64.8 & 20.5 & 53.0 & 11.0 \\
\textbf{2} & \checkmark &  &  &  & 66.6 & 26.7 & 31.2 & 25.5 & 71.3 & 42.5 & 57.0 & 18.1 \\
\textbf{3} & \checkmark & \checkmark &  &  & 69.0 & 29.1 & 33.0 & 27.1 & 74.5 & 48.4 & 59.3 & 24.1 \\
\textbf{4} & \checkmark &  & \checkmark &  & 69.4 & 29.6 & 34.0 & 28.2 & 74.4 & 48.9 & 60.2 & 24.0 \\
\textbf{5} & \checkmark & \checkmark & \checkmark &  & 70.8 & 31.4 & 34.5 & 29.4 & 75.2 & 50.1 & 60.9 & 25.8 \\
\textbf{6} & \checkmark & \checkmark & \checkmark & \checkmark & 72.0 & 32.1 & 36.0 & 30.5 & 76.0 & 50.8 & 62.0 & 26.1 \\ \hline
\end{tabular}
}
}
}
\caption{Ablation study on different components: Graph transformer (GT), hierarchical structural motifs (HSM), gait collaborative motifs (GCM), and combinatorial skeleton prototype learning (CSP). \checkmark indicates using the component.}
\label{ablation_results}
\end{table}

\subsection{Ablation Study}
We conduct ablation study to evaluate the effectiveness and contribution of each component in our approach. 
As shown in Table \ref{ablation_results}, we adopt the direct prototype learning (DP) of skeleton sequences as the baseline (ID = 1) and include GT with direct graph prototype learning  \cite{rao2023transg} (ID = 2) for comparison.
In contrast to DP or GT without employing relation learning or graph motifs, integrating motifs HSM or GCM into the body-joint relation learning obtains significantly higher mAP ($1.6$-$28.4\%$) and Rank-1 accuracy ($1.8$-$31.7\%$) on different datasets, while combining them (MGT) (ID = 5) further improves the overall performance.
This demonstrates the effectiveness of both HSM and GCM, as they can function individually or be compatibly combined to capture more discriminative relational features from structural and gait aspects for person re-ID.
Furthermore, incorporating combinatorial skeleton prototype learning (CSP) into MGT (ID = 6) consistently achieves higher results by up to $1.5\%$ for Rank-1 accuracy and $1.1\%$ for mAP on all datasets. This verifies the ability of CSP to utilize spatially-temporally combined sub-skeleton and sub-tracklet representations to enhance the capture of key skeleton patterns and class-related semantics for person re-ID.

\begin{table}[t]
\centering
\scalebox{0.7}{
\renewcommand\arraystretch{1.3}{
\setlength{\tabcolsep}{0.6mm}{
\begin{tabular}{ll|rrrrrrrrrr}
\hline
\multicolumn{2}{l|}{\textbf{Probe-Gallery}} & \multicolumn{2}{c}{\textbf{N-N}} & \multicolumn{2}{c}{\textbf{B-B}} & \multicolumn{2}{c}{\textbf{C-C}} & \multicolumn{2}{c}{\textbf{C-N}} & \multicolumn{2}{c}{\textbf{B-N}} \\ \hline
\multicolumn{2}{l|}{\textbf{Methods}} & \textbf{mAP} & \textbf{R$_{1}$} & \textbf{mAP} & \textbf{R$_{1}$} & \textbf{mAP} & \textbf{R$_{1}$} & \textbf{mAP} & \textbf{R$_{1}$} & \textbf{mAP} & \textbf{R$_{1}$} \\ \hline
\multicolumn{1}{l|}{\multirow{6}{*}{\textbf{A.}}} & LMNN \shortcite{weinberger2009distance} & — & 3.9 & — & 18.3 & — & 17.4 & — & 11.6 & — & 23.1 \\
\multicolumn{1}{l|}{} & ITML \shortcite{davis2007information} & — & 7.5 & — & 19.5 & — & 20.1 & — & 10.3 & — & 21.8 \\
\multicolumn{1}{l|}{} & ELF \shortcite{gray2008viewpoint} & — & 12.3 & — & 5.8 & — & 19.9 & — & 5.6 & — & 17.1 \\
\multicolumn{1}{l|}{} & SDALF \shortcite{farenzena2010person} & — & 4.9 & — & 10.2 & — & 16.7 & — & 11.6 & — & 22.9 \\
\multicolumn{1}{l|}{} & MLR (Features) & — & 13.6 & — & 13.6 & — & 13.5 & — & 9.7 & — & 14.7 \\
\multicolumn{1}{l|}{} & MLR (Scores) \shortcite{liu2015enhancing} & — & 16.3 & — & 18.9 & — & 25.4 & — & 20.3 & — & 31.8 \\ \hline
\multicolumn{1}{l|}{\multirow{8}{*}{\textbf{S.}}} & AGE \shortcite{rao2020self} & 3.5 & 20.8 & 9.8 & 37.1 & 9.6 & 35.5 & 3.0 & 14.6 & 3.9 & 32.4 \\
\multicolumn{1}{l|}{} & SM-SGE \shortcite{rao2021sm} & 6.6 & 50.2 & 9.3 & 26.6 & 9.7 & 27.2 & 3.0 & 10.6 & 3.5 & 16.6 \\
\multicolumn{1}{l|}{} & SPC-MGR \shortcite{rao2022skeleton} & 9.1 & 71.2 & 11.4 & 44.3 & 11.8 & 48.3 & 4.3 & 22.4 & 4.6 & 28.9 \\
\multicolumn{1}{l|}{} & SGELA \shortcite{rao2021self} & 9.8 & 71.8 & 16.5 & 48.1 & 7.1 & 51.2 & 4.7 & 15.9 & 6.7 & 36.4 \\
\multicolumn{1}{l|}{} & SimMC \shortcite{rao2022simmc} & 10.8 & 84.8 & 16.5 & 69.1 & 15.7 & 68.0 & 5.4 & 25.6 & 7.1 & 42.0 \\
\multicolumn{1}{l|}{} & TranSG \shortcite{rao2023transg} & 13.1 & 78.5 & 17.9 & 67.1 & 15.7 & 65.6 & 6.7 & 23.0 & 8.6 & 44.1 \\
\multicolumn{1}{l|}{} & Hi-MPC \shortcite{rao2024hierarchical} & 11.2 & 85.5 & 17.0 & 71.2 & 14.1 & 70.2 & 4.9 & \textbf{27.2} & 7.5 & 50.1 \\
\multicolumn{1}{l|}{} & \textbf{MoCos (Ours)} & \textbf{16.1} & \textbf{87.9} & \textbf{18.9} & \textbf{73.6} & \textbf{18.1} & \textbf{72.1} & \textbf{7.3} & 26.5 & \textbf{9.8} & \textbf{50.6} \\ \hline
\end{tabular}
}
}
}
\caption{Person re-ID performance comparison with \textbf{A}ppearance-based (\textbf{A.}) or \textbf{S}keleton-based (\textbf{S.}) methods on CASIA-B. ``\textbf{C-N}'' denotes using ``\textbf{C}lothes'' probe set and ``\textbf{N}ormal'' gallery set. ``—'' indicates no published result. }
\label{CASIA_B_results}
\end{table}

\section{Further Analysis}
\label{discussions}

\quad\textbf{Application to RGB-estimated Scenarios.}
To verify the generality of MoCos on RGB-estimated skeletons, we extract skeleton data with pre-trained pose estimation models \cite{cao2019openpose,chen20173d} from RGB videos instead of depth sensors.
The results in Table \ref{CASIA_B_results} show that our model not only achieves superior performance to most existing state-of-the-art skeleton-based models, but also outperforms many representative established appearance-based methods that rely on RGB-based features ($e.g.,$ silhouettes) or/and visual metric learning \cite{liu2015enhancing,farenzena2010person}.
This verifies the generality and higher effectiveness of MoCos to learn discriminative patterns from estimated skeletons, and demonstrates its potential for person re-ID under large-scale RGB-based scenarios.

\begin{table}[t]
\scalebox{0.7}{
\renewcommand\arraystretch{1.3}{
\setlength{\tabcolsep}{1.3mm}{
\begin{tabular}{llrrrrrrrr}
\hline
\multirow{2}{*}{\textbf{Scales}} & \multirow{2}{*}{\textbf{Methods}} & \multicolumn{2}{c}{\textbf{BIWI-S}} & \multicolumn{2}{c}{\textbf{BIWI-W}} & \multicolumn{2}{c}{\textbf{KS20}} & \multicolumn{2}{c}{\textbf{KGBD}} \\
                                 &                                   & \textbf{R$_{1}$}   & \textbf{mAP}   & \textbf{R$_{1}$}   & \textbf{mAP}   & \textbf{R$_{1}$}  & \textbf{mAP}  & \textbf{R$_{1}$}  & \textbf{mAP}  \\ \hline
\multirow{2}{*}{\textbf{J-Scale}}     & SM-SGE \shortcite{rao2021sm}                            & 33.0               & 10.0           & 12.9               & 14.9           & 44.7              & 10.2          & 40.2              & 4.3           \\
                                 & MoCos (Ours)                      & \textbf{72.0}      & \textbf{32.1}  & \textbf{36.0}      & \textbf{30.5}  & \textbf{76.0}     & \textbf{50.8} & \textbf{62.0}     & \textbf{26.1} \\ \hline
\multirow{2}{*}{\textbf{P-Scale}}      & SM-SGE                            & 32.8               & 11.1           & 14.5               & 16.5           & 43.2              & 9.8           & 33.0              & 4.1           \\
                                 & MoCos (Ours)                      & \textbf{38.3}      & \textbf{14.7}  & \textbf{20.7}      & \textbf{19.0}  & \textbf{49.0}     & \textbf{15.7} & \textbf{35.9}     & \textbf{4.7}  \\ \hline
\multirow{2}{*}{\textbf{B-Scale}}      & SM-SGE                            & 27.5               & 10.0           & 12.6               & 13.8           & 37.3              & 9.3           & \textbf{31.5}     & \textbf{4.4}  \\
                                 & MoCos (Ours)                      & \textbf{35.6}      & \textbf{12.5}  & \textbf{18.4}      & \textbf{16.9}  & \textbf{41.1}     & \textbf{13.5} & 30.6              & \textbf{4.4}  \\ \hline
\end{tabular}
}
}
}
\caption{Performance of MoCos on \textbf{J}oint (\textbf{J}), \textbf{P}art (\textbf{P}) or \textbf{B}ody (\textbf{B}) scale graph modeling with $J$, 10, and 5 nodes.}
\label{graph_scale}
\end{table}

\textbf{Evaluation on Different-Scale Skeleton Graphs.}
 We construct different-scale graphs \cite{rao2021sm} for MoCos learning to evaluate its performance under varying graph modeling. 
As presented in Table \ref{graph_scale}, compared with the state-of-the-art multi-scale graph method SM-SGE \cite{rao2021sm}, our model achieves better performance in most cases of both original and higher level skeleton representations ($e.g.,$ part-scale skeleton graphs). Such results suggest the compatibility of the proposed motif guided graph transformer (MGT) with different-scale graph modeling, and also justify its stronger capability to learn more effective graph features and semantics at different levels for person re-ID.

\begin{table}[t]
\centering
\scalebox{0.7}{
\renewcommand\arraystretch{1.3}{
\setlength{\tabcolsep}{1.75mm}{
\begin{tabular}{lrrrrrrrr}
\hline
\multirow{2}{*}{\textbf{Methods}} & \multicolumn{2}{c}{\textbf{BIWI-S}} & \multicolumn{2}{c}{\textbf{BIWI-W}} & \multicolumn{2}{c}{\textbf{KS20}} & \multicolumn{2}{c}{\textbf{KGBD}} \\
 & \textbf{mAP} & \textbf{R$_{1}$} & \textbf{mAP} & \textbf{R$_{1}$} & \textbf{mAP} & \textbf{R$_{1}$} & \textbf{mAP} & \textbf{R$_{1}$} \\ \hline
SPC-MGR \shortcite{rao2022skeleton} & 16.0 & 34.1 & 19.4 & 18.9 & 21.7 & 59.0 & 6.9 & 40.8 \\
SPC-MGR + MoCos & \textbf{16.3} & \textbf{42.8} & \textbf{20.1} & \textbf{23.6} & \textbf{23.6} & \textbf{65.4} & \textbf{8.2} & \textbf{43.2} \\ \hline
SimMC \shortcite{rao2022simmc} & 12.3 & 41.7 & 19.9 & 24.5 & 22.3 & 66.4 & 11.7 & \textbf{54.9} \\
SimMC + MoCos & \textbf{16.0} & \textbf{55.4} & \textbf{22.4} & \textbf{25.9} & \textbf{23.9} & \textbf{67.2} & \textbf{12.1} & 54.6 \\ \hline
\end{tabular}
}
}
}
\caption{Performance of our approach when applied to different unsupervised paradigms using \textit{unlabeled} skeletons.}
\label{unsupervised_models}
\end{table}

\begin{figure}[t]
\centering
	\subcaptionbox{$t$-SNE Visualization}{\scalebox{0.25}{\includegraphics[]{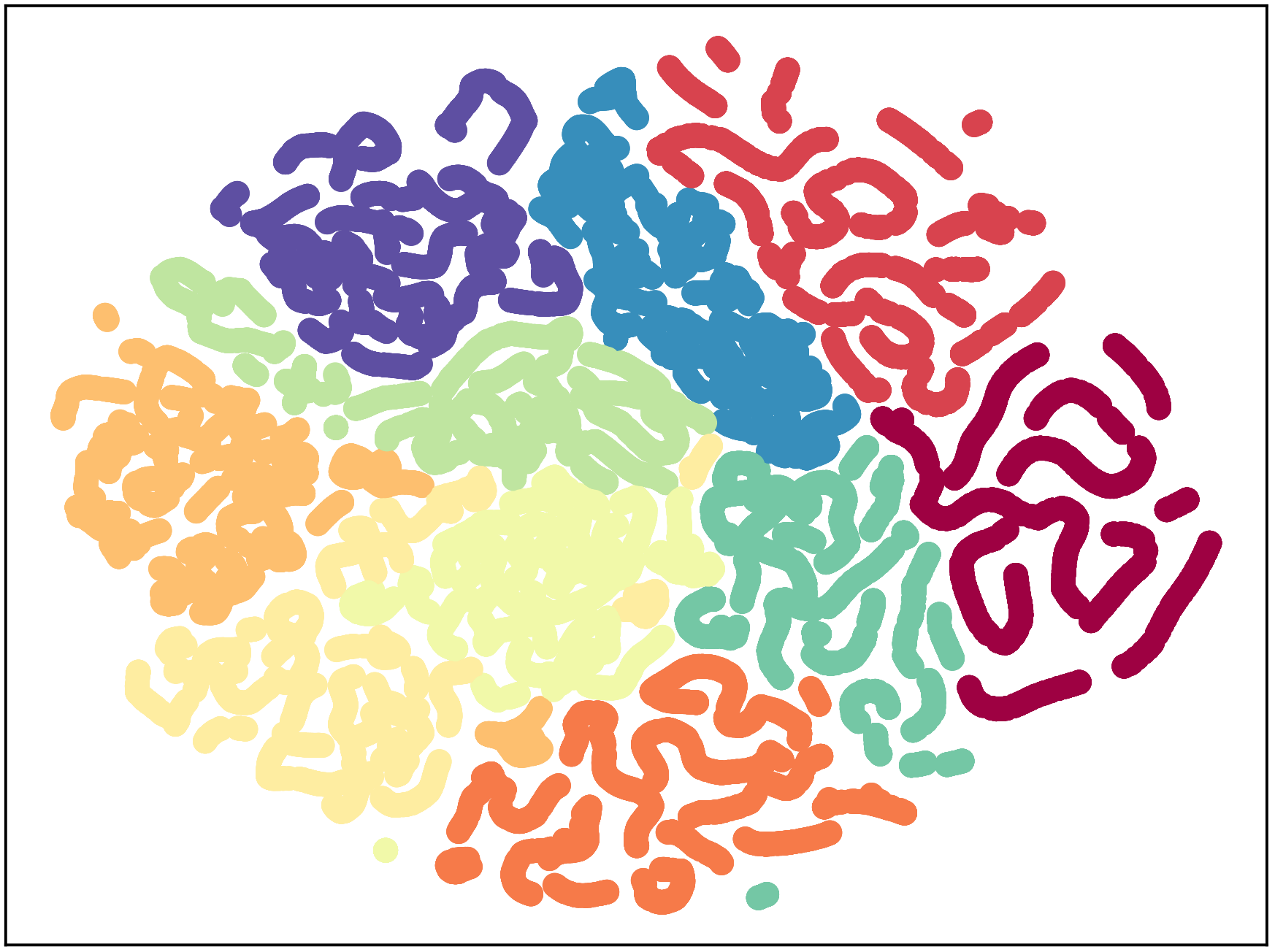}} \scalebox{0.25}{\includegraphics[]{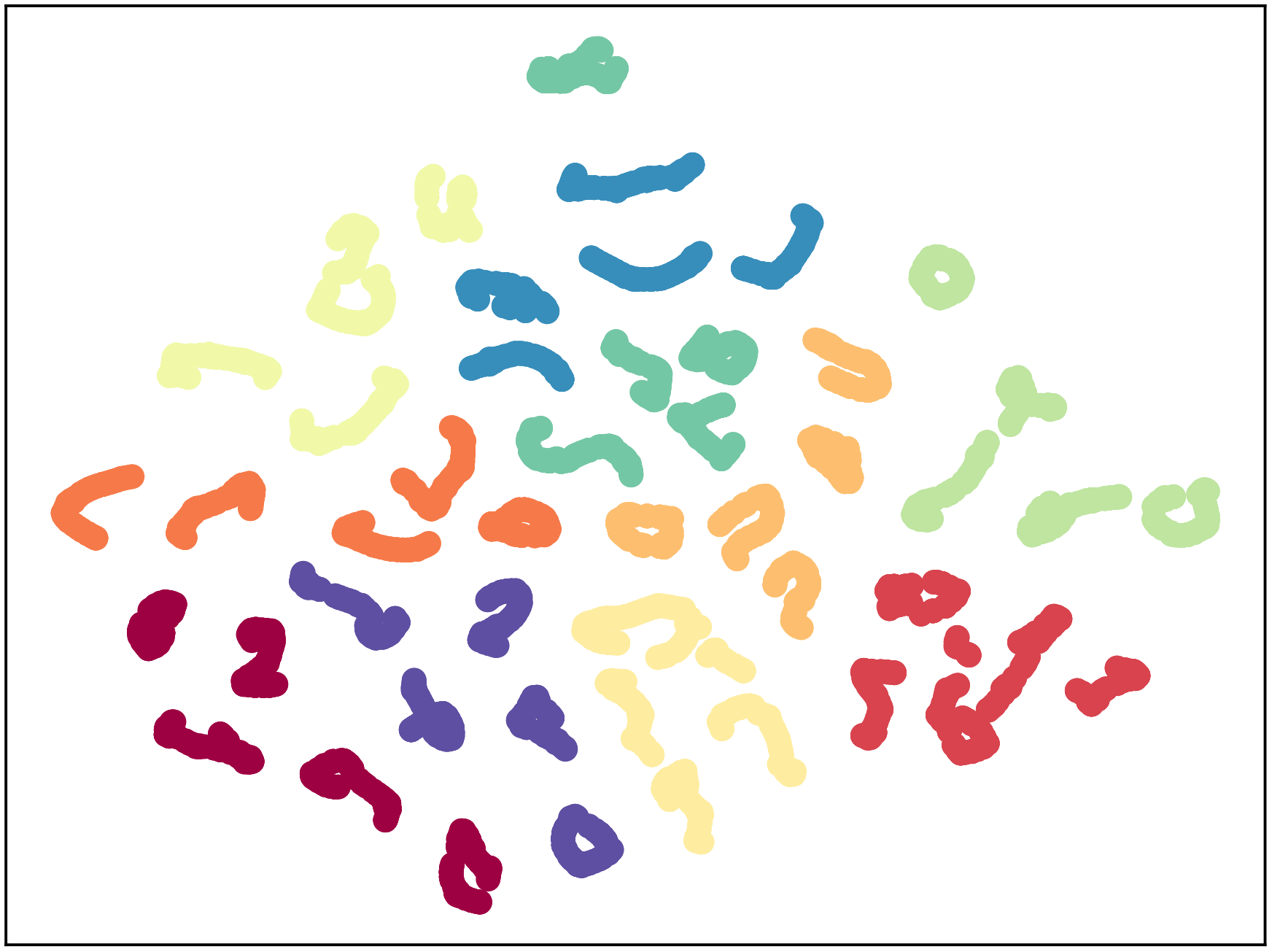}}}
     \ \ 
     \subcaptionbox{Relation Comparison}{\scalebox{0.3}{\includegraphics[]{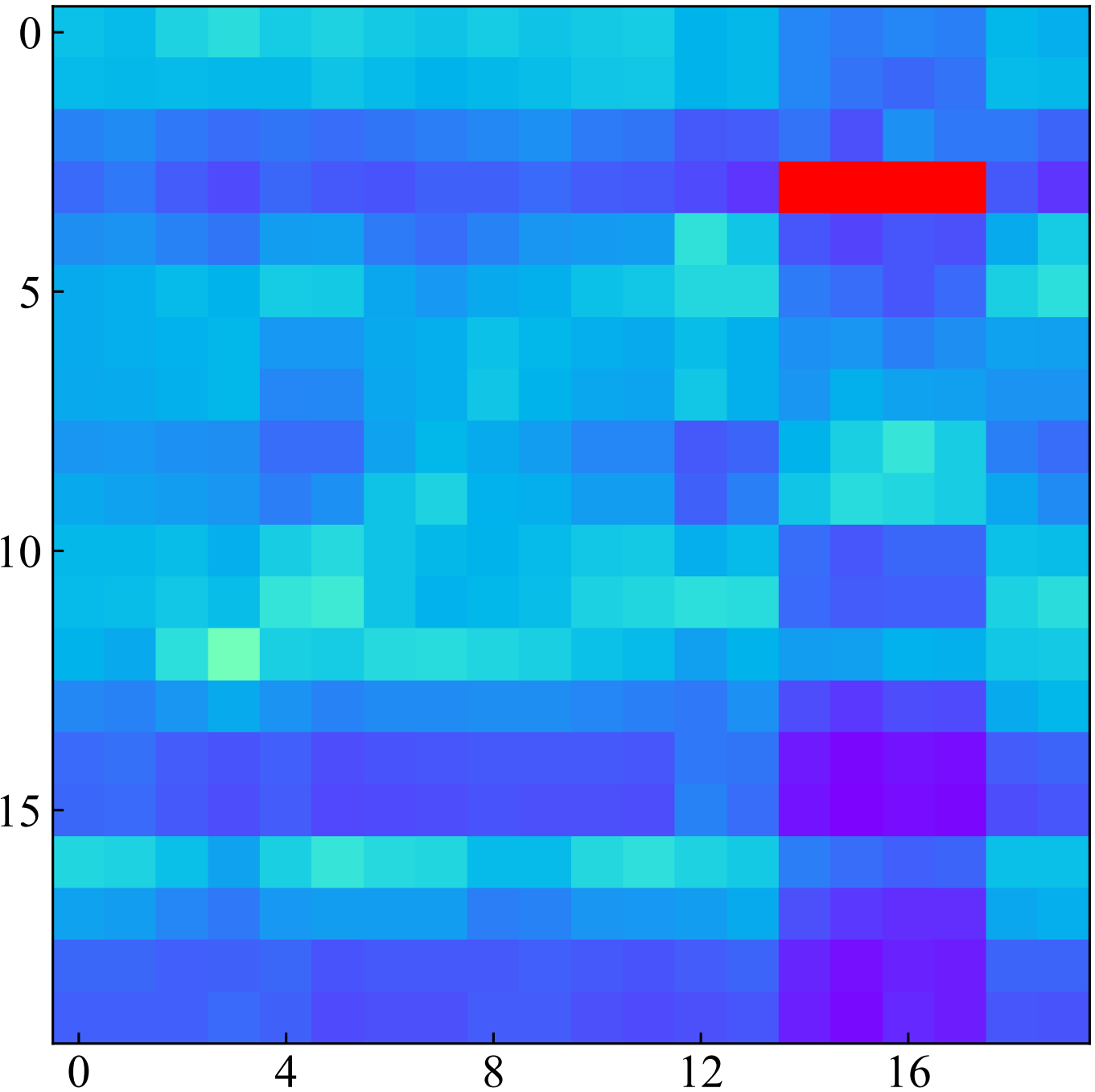}} \quad \scalebox{0.3}{\includegraphics[]{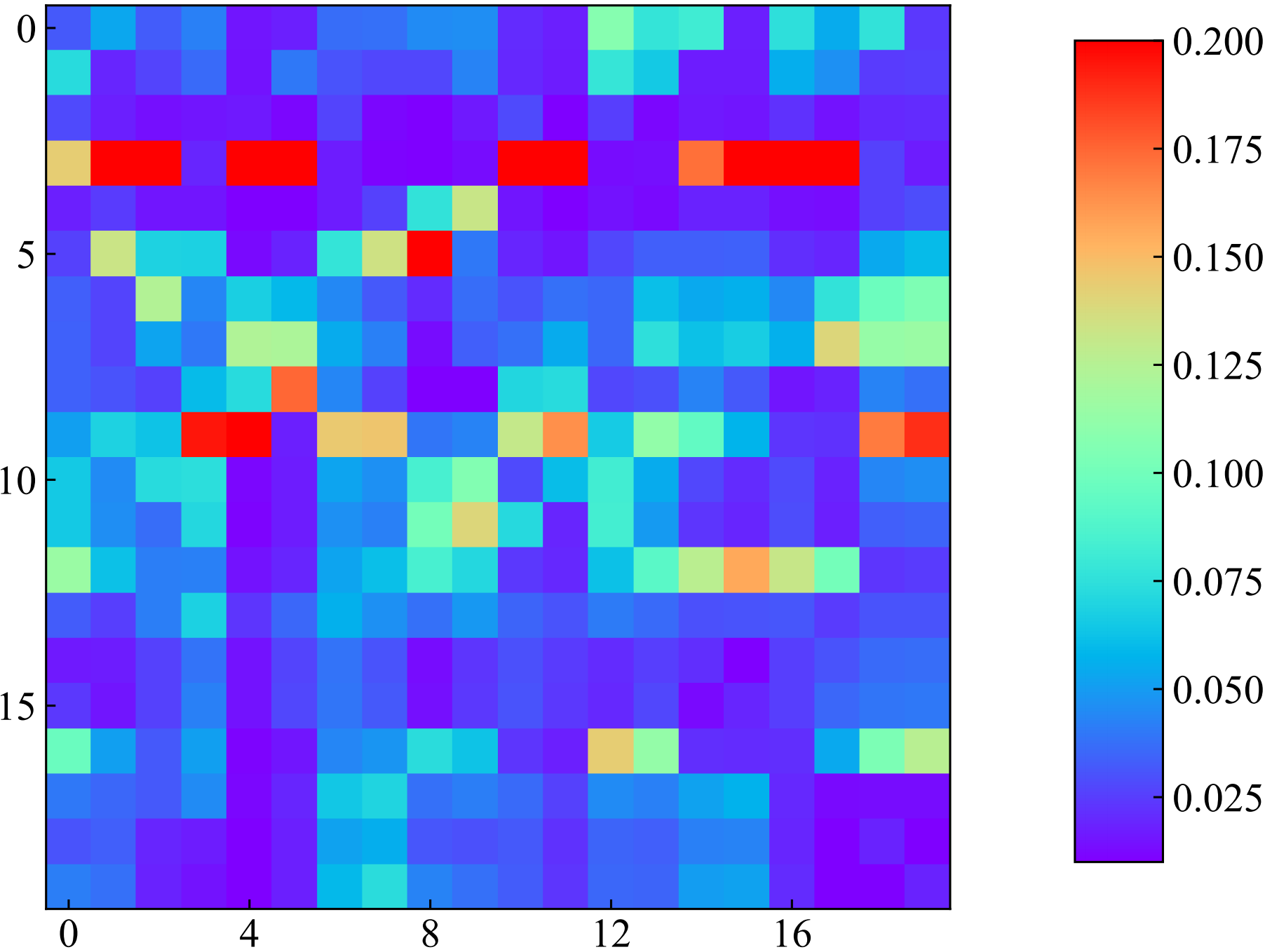}}}
\caption{(a) $t$-SNE visualization of features for the first ten classes in IAS and KS20. Different colors indicates different classes. (b) Visualization of mean relation values inferred by non-motif method \cite{rao2023transg} (Left) and our MoCos (Right) on the same value scale and testing skeletons.}
\label{visualization}
\end{figure}

\textbf{Transfer to Unsupervised Paradigms.}
 As shown in Table \ref{unsupervised_models}, our MGT and CSP (denoted as “+ MoCos”) can be transferred for \textit{unlabeled} skeleton relation and prototype learning, which effectively boosts performance of different unsupervised non-graph and non-transformer models \cite{rao2022skeleton,rao2022simmc} in most cases. This validates generality and scalability of MoCos, which can be potentially applied to more general scenarios without labels.

\textbf{Feature and Relation Visualization.}
The $t$-SNE visualization \cite{van2008visualizing} in Fig. \ref{visualization} (a) shows the evident inter-class separation of the learned features, suggesting the effectiveness of our approach to capture useful class-related semantics on different datasets.  
We also visualize the mean relations of ($J=20$) joints inferred from MoCos in Fig. \ref{visualization} (b), and the results imply that our motif-guided approach could capture richer and more salient joint correlations than TranSG \cite{rao2023transg} that solely uses full-relation learning without motifs.
More empirical and theoretical analyses are provided in Appendix I and II.

\section{Conclusion}
In this paper, we propose MoCos to perform motif-guided joint relation learning and combinatorial skeleton prototype learning for person re-ID. We design the motif guided graph transformer (MGT) that incorporates hierarchical structural motifs and gait collaborative motifs to capture key relations within multi-order body joints' structure and gait-related limbs. The combinatorial skeleton prototype learning (CSP) is proposed to contrast randomly-combined sub-skeleton and sub-tracklet graph features with skeleton prototypes to learn class-related semantics and discriminative representations. Our approach outperforms existing state-of-the-art models, and can be generally applied to various scenarios.

\section{Acknowledgements}
This research is supported by the National Research Foundation, Singapore under its AI Singapore Programme (AISG Award No: AISG2-PhD/2022-01-034[T]).


\bibliography{main}

\end{document}